\documentclass{article}
\usepackage[utf8]{inputenc}
\usepackage{graphicx}
\usepackage{multirow}
\usepackage{url}
\usepackage{hyperref}
\usepackage[table]{xcolor}
\usepackage{arxiv}
\usepackage{siunitx}
\usepackage{amsmath}

\usepackage{hyperref}
\hypersetup{
pdftitle={A Comparison between Frame-based and Event-based Cameras for Flapping-Wing Robot Perception},
pdfauthor={R. Tapia, J.P. Rodríguez-Gómez, J.A. Sanchez-Diaz, F.J. Gañán, I.G. Rodríguez, J. Luna-Santamaria, J.R. Martínez-de Dios, and A. Ollero},
pdfsubject={2023 IEEE/RSJ International Conference on Intelligent Robots and Systems}
}

\title{A Comparison between Frame-based and Event-based Cameras for Flapping-Wing Robot Perception}

\author{
\href{https://orcid.org/0000-0002-4435-5466}{\includegraphics[scale=0.06]{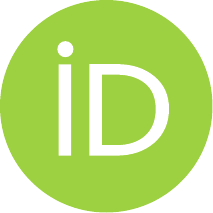}\hspace{1mm}Raul Tapia}\\
	GRVC Robotics Lab.\\
	Universidad de Sevilla\\
	\hphantom{00}\texttt{raultapia@us.es}\hphantom{00}\\
	\And
\href{https://orcid.org/0000-0001-7628-1660}{\includegraphics[scale=0.06]{orcid.pdf}\hspace{1mm}Juan Pablo Rodríguez-Gómez}\\
	GRVC Robotics Lab.\\
	Universidad de Sevilla\\
	\hphantom{00}\texttt{jrodriguezg@us.es}\hphantom{00}\\
	\And
\href{https://orcid.org/0009-0008-3022-7499}{\includegraphics[scale=0.06]{orcid.pdf}\hspace{1mm}Juan Antonio Sanchez-Diaz}\\
	GRVC Robotics Lab.\\
	Universidad de Sevilla\\
	\hphantom{00}\texttt{jsanchez22@us.es}\hphantom{00}\\
	\And
\href{https://orcid.org/0000-0003-1599-7424}{\includegraphics[scale=0.06]{orcid.pdf}\hspace{1mm}Francisco Javier Gañán}\\
	GRVC Robotics Lab.\\
	Universidad de Sevilla\\
	\hphantom{00}\texttt{fgannan@us.es}\hphantom{00}\\
	\And
\href{https://orcid.org/0009-0009-7222-5835}{\includegraphics[scale=0.06]{orcid.pdf}\hspace{1mm}Iván Gutierrez Rodríguez}\\
	GRVC Robotics Lab.\\
	Universidad de Sevilla\\
	\hphantom{00}\texttt{igrodriguez@us.es}\hphantom{00}\\
	\And
\href{https://orcid.org/0009-0009-5739-1894}{\includegraphics[scale=0.06]{orcid.pdf}\hspace{1mm}Javier Luna-Santamaria}\\
	GRVC Robotics Lab.\\
	Universidad de Sevilla\\
	\hphantom{00}\texttt{javierluna@us.es}\hphantom{00}\\
	\And
\href{https://orcid.org/0000-0001-9431-7831}{\includegraphics[scale=0.06]{orcid.pdf}\hspace{1mm}José Ramiro Martínez-de Dios} \\
	GRVC Robotics Lab.\\
	Universidad de Sevilla\\
	\texttt{jdedios@us.es}\\
	\And
\href{https://orcid.org/0000-0003-2155-2472}{\includegraphics[scale=0.06]{orcid.pdf}\hspace{1mm}Anibal Ollero} \\
	GRVC Robotics Lab.\\
	Universidad de Sevilla\\
	\texttt{aollero@us.es}\\
}

\paperurl{https://raultapia.com/preprints/0004/redirect}
\journal{2023 IEEE/RSJ International Conference on Intelligent Robots and Systems}
\editorial{IEEE}

\begin{document}

\maketitle

\begin{abstract}
Perception systems for ornithopters face severe challenges. The harsh vibrations and abrupt movements caused during flapping are prone to produce motion blur and strong lighting condition changes. Their strict restrictions in weight, size, and energy consumption also limit the type and number of sensors to mount onboard. Lightweight traditional cameras have become a standard off-the-shelf solution in many flapping-wing designs. However, bioinspired event cameras are a promising solution for ornithopter perception due to their microsecond temporal resolution, high dynamic range, and low power consumption. This paper presents an experimental comparison between frame-based and an event-based camera. Both technologies are analyzed considering the particular flapping-wing robot specifications and also experimentally analyzing the performance of well-known vision algorithms with data recorded onboard a flapping-wing robot. Our results suggest event cameras as the most suitable sensors for ornithopters. Nevertheless, they also evidence the open challenges for event-based vision on board flapping-wing robots.
\end{abstract}

\section{Introduction}
\label{sec:intro}
Flapping-wing robots --also known as ornithopters-- are bioinspired aerial platforms that generate lift and thrust through oscillating flapping wings. They have high maneuverability without using fast rotating propellers, and combine gliding and flapping modes to reduce power consumption \cite{croon2020flapping}. Besides,  they are less dangerous and more robust against collisions than multirotors and fixed-wing platforms. Due to their potentialities for a wide range of applications \cite{zufferey2022how}, they have attracted significant R\&D interest in the last years \cite{ollero2022past}.

\begin{figure}
    \centering
    \includegraphics[width=0.8\linewidth,page=10]{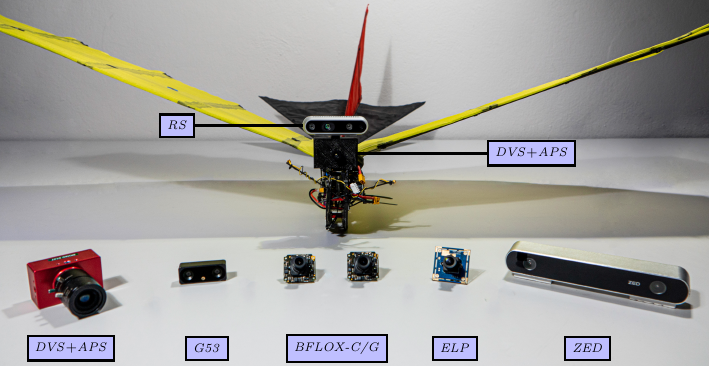}
    \caption{\textit{E-Flap} flapping-wing robot used in the experiments and the set of cameras evaluated in this work.}
    \label{fig:intro}
\end{figure}

We are interested in developing online onboard perception methods for flapping-wing robots. However, they involve significantly stronger perception challenges than other aerial platforms. Their strict payload, weight distribution, and energy consumption constrain the number and type of sensors to mount onboard. In addition, their fast maneuvers, vibrations, and abrupt tilt changes caused by flapping strokes may perturb perception measurements (e.g., motion blur in intensity images) \cite{croon2012appearance} \cite{gomez2019towards}. These constraints discourage the use of some commonly-applied sensors in aerial robotics such as LiDARs, particularly when the payload is a hard restriction. Cameras emulate the vision mechanism of most animals. They are small and lightweight, and there are plenty of available vision algorithms for many robot perception problems. The first perception systems for ornithopters were based on frame cameras \cite{croon2009design,garcia2009optical} and optical-flow-based sensors \cite{duhamel2012altitude,duhamel2013biologically}. Recent works use event cameras \cite{gomez2021why,rodriguez2022free}, as these bioinspired sensors provide high responsiveness, dynamic range, and robustness against motion blur \cite{gallego2020event}. Although there are different approaches and opinions, no work has reported an in-depth analysis of frame-based and event-based cameras for ornithopter robot perception.

This paper presents an experimental comparison between different frame cameras and an event camera for online onboard ornithopter perception, see Figure \ref{fig:intro}. It analyzes them using criteria based on the requirements and constraints of the platform (weight, size, and power consumption), the nature and challenges of flapping-wing flight (dynamic range, motion blur), and compares the performance of several well-known framed-based and event-based vision algorithms. The analysis includes experiments in test-benches that emulate flapping conditions and also flights on board an ornithopter. The main contributions of the paper are: i) qualitative and quantitative analyses of different visual sensors based on ornithopter platform and flight requirements; and ii) a comparison of different algorithms for corner detection, visual-inertial odometry, line detection, and human detection on images and events collected in ornithopter flights.

This paper is structured as follows. Section \ref{sec:soa} summarizes the main works on ornithopter perception systems. The main requirements and challenges are described in Section \ref{sec:ornithopter_perception}. Sections \ref{sec:platform} and \ref{sec:flight} compare event-based and traditional cameras in terms of the flapping-wing platform specifications and its flight requirements. Section \ref{sec:application} discusses the convenience of both types of cameras in different computer vision problems. Section \ref{sec:discussion} provides a final discussion, and Section \ref{sec:conclusion} summarizes the conclusions and future work.

\section{Related Work}
\label{sec:soa}
Online perception on board flapping-wing robots involves strong challenges. In fact, ornithopter control and guidance methods have traditionally relied on external sensors such as motion capture systems \cite{ma2013controlled,maldonado2020adaptive}. In the last years, the advances in size and weight reduction of vision sensors have facilitated the integration of online perception systems on ornithopters \cite{croon2009design}. A mechanical camera stabilizer is described in \cite{pan2020development}. It compensates for the vibrations caused by flapping while reducing the robot payload. An onboard vision-based target tracking system for small ornithopters is proposed in \cite{ryu2016autonomous}. It is based on a lightweight camera that transmits video to a base station where it is processed. The flapping-wing perception dataset in \cite{rodriguez2021griffin} includes measurements recorded from an event camera and a traditional camera to pave the way for the development of ornithopter perception methods. Although earlier works present off-board perception approaches \cite{croon2009design,garcia2009optical,duhamel2012altitude,duhamel2013biologically,croon2012appearance}, few authors have integrated onboard processing in ornithopters. One of the first was an obstacle avoidance method which used a lightweight rolling shutter stereo setup to mitigate the flapping motion, and includes an embedded CPU to detect statics obstacles using dense stereo matching \cite{wagter2014autonomous} or sparse stereo matching \cite{tijmons2017obstacle}. The work in \cite{rodriguez2022free} presents an event-based dynamic obstacle avoidance method for ornithopters. The method detects moving obstacles and triggers evasive actions controlling the ornithopter tail deflections. The authors in \cite{gomez2021why} present an event-based ornithopter guidance method. The algorithm tracks line pattern references from events and feeds a visual servoing controller to guide the robot towards the goal. The above works depict the relevance of visual sensors for ornithopter perception. However, they use different types of vision sensors (frame-based monocular and stereo sensors, and also event cameras) and argue on the goodness of their sensor for flapping-wing robot perception, showing a lack of consensus on the topic.

Despite the increasing interest attracted by event cameras, few studies have experimentally compared frame-based and event cameras. The authors in \cite{holesovsky2020practical,holesovsky2021experimental} use two test bench setups to experimentally compare an ATIS HVGA Gen3 event camera, a DVS240 event camera, and two high-speed global-shutter frame cameras. The results include sampling and detection rates as functions of the motion speed and the scene illuminance, position estimation errors, and pixel latency. The authors conclude that event cameras outperform frame cameras in bandwidth efficiency, although they may be limited by pixel latency and/or readout bandwidth, especially in highly cluttered scenes. Although event cameras bandwidth may entail a limitation in some applications, the use of novel neuromorphic processors such as spiking neural networks becomes a potential solution \cite{paredes2020unsupervised} \cite{gehrig2020event}. The work in \cite{barrios2018movement} presents a comparison between a GENIE M640 CCD camera and a noncommercial event CMOS camera to control a two-axis planar robot for object tracking. The results show that, using the event camera, the robot can follow the target faster, more accurately, and more stable under light changes.

Some works have reported theoretical comparisons of event and frame cameras for specific tasks. The work \cite{censi2015power} presents a formal comparison of different sensor families (focusing on traditional CCD/CMOS versus neuromorphic sensors) through a power-performance curve. It concludes that, depending on the task, different kinds of sensor can dominate the others in different ranges of sensing power. A theoretical methodology to analyze the performance of event and frame-based cameras is presented in \cite{cox2020analysis}. It proposes an implementation example using system-level models, and surrogates performance metrics for a target recognition application. Other works have compared them from a data processing perspective. The work in \cite{farabet2012comparison} analyzes frame-based convolutional neural networks and frame-free spiking neural networks for object recognition, including examples of implementation using VLSI chips and FPGAs. It compares them in terms of computational speed, scalability, multiplexing, and signal representation. Works \cite{rebecq2019events, rebecq2021high} present an image reconstruction method based on a recurrent neural network and compare the quality of the reconstructed images, validating their results with visual-inertial odometry and object classification algorithms. Although the above works propose different analysis of the performance of frame-based and event-based cameras, the experimental comparison of those sensors onboard aerial robots remains a neglected area. To the best of author's knowledge, this is the first work experimentally analyzing and comparing those sensor families for flapping-wing robot perception.

\newcolumntype{C}{>{\vspace{0.08mm}}c}
\newcommand\cincludegraphics[2][]{\raisebox{-0.3\height}{\includegraphics[#1]{#2}}}

\begin{table*}
\centering
\renewcommand{\arraystretch}{1.2}
\scalebox{0.75}{
\begin{tabular}{CCCCCCCCCC}
\rowcolor{blue!25}
\textbf{Company} & \textbf{Name} & \textbf{Alias} & \textbf{Type} & \textbf{Ch.} & \textbf{Resolution} & \textbf{Dimensions (mm)} & \textbf{Depth (m)} & \textbf{FPS} & \textbf{Weight (g)} \\
iniVation & DAVIS346 & \textit{DVS}, \textit{APS} & EVENT & 1 & \begin{tabular}[c]{@{}c@{}}346$\times$260\end{tabular} & 40$\times$60$\times$25 &  - & APS:30 & 100 \\
eCapture & G53 & \textit{G53} & STEREO & 3 & 640$\times$400 & 50$\times$14.9$\times$20 & 0.15 - 2 & 30 & 100 \\
Intel & Realsense D435 & \textit{RS} & STEREO & 3 & 1280$\times$720 & 90$\times$25$\times$25 & 0.3 - 3 & \begin{tabular}[c]{@{}c@{}}30\end{tabular} & 340 \\
ELP & Mini720p & \textit{ELP} & MONO & 3 & 1280$\times$720 & 39$\times$39$\times$20 &  - & 30 & 17\\
StereoLabs & ZED & \textit{ZED} & STEREO & 3 & 1280$\times$720 & 175$\times$30$\times$33 & 0.5 - 25 & 60 & 170 \\
Matrix Vision & mvBlueFOX MLC200wC & \textit{BFOX-C} & MONO & 3 & 752$\times$480 & 35$\times$33$\times$25 &  - & 90 & 10 \\
Matrix Vision & mvBlueFOX MLC200wG & \textit{BFOX-G} & MONO & 1 & 752$\times$480 & 35$\times$33$\times$25 &  - & 90 & 10 \\
\end{tabular}
}
\caption{List of vision sensors used in the experimental comparisons throughout this paper. In those cases where the resolution and/or FPS of a sensor are configurable, the table presents the selected values used in the analyses.}
\label{tab:cameras}
\end{table*}

\begin{figure*}
    \centering
    \includegraphics[width=\linewidth,page=6]{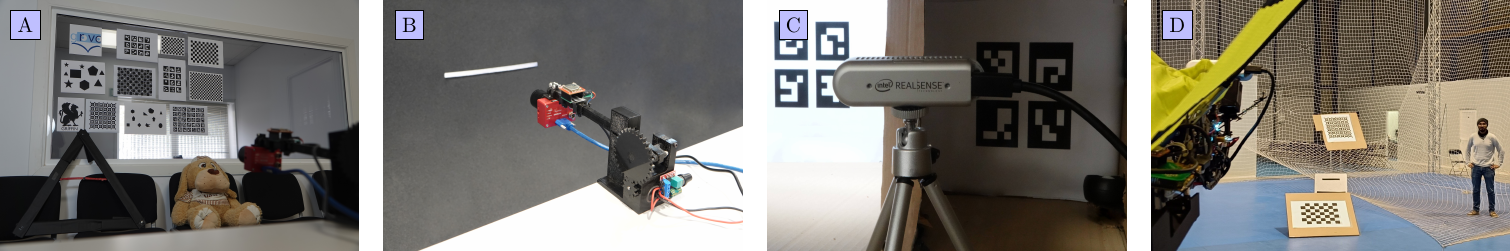}
    \caption{Setups of the three test benches (a,b,c) and the ornithopter flight experiments (d). Setups a) and b) include a mechanism to emulate the pitch camera motion due to flapping strokes.}
    \label{fig:setups}
\end{figure*}

\section{Ornithopter Perception Challenges}
\label{sec:ornithopter_perception}
The design of perception systems for flapping-wing robots involves several constraints. First, ornithopters have strict payload and weight distribution restrictions, which impact the weight and shape of sensors, electronics, batteries, and other components. Further, their payload constraints impose severe energy consumption considerations. Flapping strokes generate mechanical vibrations and abrupt pitch changes, that can cause motion blur and strong changes in lighting conditions \cite{croon2012appearance,gomez2019towards}. Besides, they can modify the flapping frequency, affecting these constraints. Hence, the sensors for flapping-wing robots should be selected considering their responsiveness to agile motions, robustness against vibrations, weight, shape, and energy consumption, among others.

The above requirements prevent the use on ornithopters of a wide variety of sensors that are common in multirotors. The majority of 2D and 3D LiDARs are not suitable due to payload and weight distribution constraints. In addition, their scan rate is often insufficient for fast maneuvers and some of them are not suitable for outdoor perception. Solid-state LiDARs enable higher miniaturization and have higher scan rates. However, they have higher power consumption and weight than the cameras that are commonly used. Although radar miniaturization advances enable their use on aerial robots, their weight is still high for moderate-size ornithopters, which payloads are of a few hundred of grams \cite{zufferey2021design}. Also, lightweight infrared and multi-spectral cameras require large exposure times \cite{cutler2013lightweight}, causing motion blur due to ornithopter strong vibrations. Ultrasound sensors also have poor performance since the abrupt pitch changes due to flapping strokes hinder the reception of the reflected signals, causing false negatives in presence of nearby obstacles.

Vision sensors have low size and weight. They are suitable for online onboard ornithopter perception, and in fact, all existing related works selected different types of cameras as the main sensor. Despite this agreement, there are different approaches and opinions on the type of vision sensor to be used. This work intends to contribute to this discussion by presenting qualitative and quantitative comparisons between different vision sensors. A total of 7 cameras, see Table \ref{tab:cameras}, are analyzed: i) a DAVIS346 which includes a \textit{DVS} event-based sensor and an \textit{APS} frame-based sensor, ii) a widely-used RGB-D Realsense D435 (\textit{RS}), iii) a high-resolution ZED stereo pair (\textit{ZED}), iv) a lightweight G53 stereo pair (\textit{G53}), v) a lightweight low-cost 720p camera (\textit{ELP}), and vi-vii) two lightweight high frame rate mvBlueFOX cameras (\textit{BFOX-C} and \textit{BFOX-G}). They were selected due to their diverse resolutions, dimensions, weights, and FPS, and they could be potentially used for ornithopter perception. Our analysis includes the DAVIS346, the only event camera that has been mounted on an ornithopter \cite{rodriguez2021griffin}. Despite the low-resolution of \textit{DVS}, the work in \cite{gehrig2022are} suggests that low-resolution event cameras might report better performance than higher resolution event cameras under high-speed motions and low lighting, both conditions that arise on ornithopter flight.

The presented comparison has a strong experimental focus and includes analyses in three different test benches and flight experiments on board an ornithopter, see Figure \ref{fig:setups}. The vision sensors are compared using the above ornithopter requirements, which can be classified into:
\begin{itemize}
    \item \textit{Ornithopter platform requirements}, derived from their payload, size, and weight distribution constraints.
    \item \textit{Flapping-wing flight requirements}, derived from their fast motion, strong vibrations, and wide pitch changes.
    \item \textit{Application dependent requirements}, derived from the performance specifications of commonly-used computer vision algorithms for frame-based and event cameras.
\end{itemize}

\section{Ornithopter Platform Requirements}
\label{sec:platform}

\subsection{Weight and Size}
\label{subsec:weight_size}
\begin{figure}
    \centering
    \includegraphics[width=0.7\linewidth,page=1]{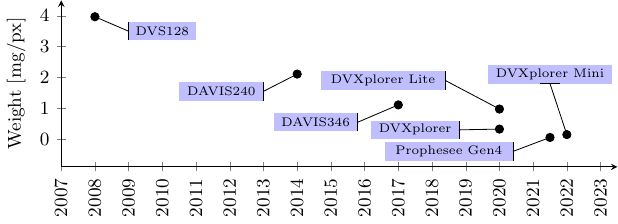}
    \includegraphics[width=0.7\linewidth,page=2]{figures.pdf}
    \caption{Evolution of weight and volume of commercial event cameras (expressed relative to the detector resolution).}
    \label{fig:weight_and_volume}
\end{figure}

Although some recent ornithopter designs have relatively higher payload \cite{zufferey2021design}, the use of small and lightweight cameras is still required. Moreover, the cameras are usually mounted at the front, significantly affecting the center of mass, which is critical for the platform maneuverability and stability. Frame cameras are generally smaller and lighter than event cameras. However, the evolution of dynamic vision technology has originated a clear trend in reducing the event cameras weight and size, see Figure \ref{fig:weight_and_volume}. In the last years the event cameras' weight and volume are becoming similar to those of frame cameras. For example, the DVXplorer Mini event camera from iniVation has a resolution of 640$\times$480, a volume of $\sim$\SI{27}{\cubic\cm}, and a weight of $\sim$\SI{20}{\gram}. These values are similar to many frame cameras, e.g., \textit{ELP}, see Table \ref{tab:cameras}, which has a volume of $\sim$\SI{30}{\cubic\cm} and a weight of $\sim$\SI{17}{\gram}. Using weight and shape criteria, event cameras are now a viable alternative to traditional cameras.

\subsection{Power Consumption}
\label{subsec:power}
Due to the limited power capacity of lightweight batteries, energy consumption becomes a side-effect of the strict payload of ornithopters \cite{tapia2023experimental}. In this section, the electric consumption of the cameras in Table \ref{tab:cameras} is experimentally analyzed using the setup shown in Figure \ref{fig:setups}-a. The cameras were placed on a motorized mechanism that emulates the flapping strokes by varying the pitch angle $\pm$ \SI{30}{\degree} and moving linearly \SI{2.5}{\cm} back and forth. The setup includes a VectorNav VN-200 inertial navigation system to measure pitch rate and an INA219 power monitor to measure electric consumption. The data output from each camera was recorded during \SI{40}{\second} while the pitch rate systematically increased from \SI{0}{\hertz} to \SI{6}{\hertz}. The typical flapping frequency of \textit{E-Flap} ornithopter is \SI{3.5}{\hertz} \cite{zufferey2021design}. In fact, in our experiments in Section \ref{sec:application} the mean flapping frequency was $\mu_f = $ \SI{4.31}{\hertz}, with standard deviation $\sigma_f = $ \SI{0.833}{\hertz}. The power consumption of event cameras depends on the number of triggered events. Hence, the experiment was repeated in three different scenarios: \textit{static}, \textit{low-dynamic} (some objects in the scene moved slowly), and \textit{high-dynamic scenarios} (some objects moved fast).

\begin{figure}
    \centering
    \includegraphics[width=0.8\linewidth,page=3]{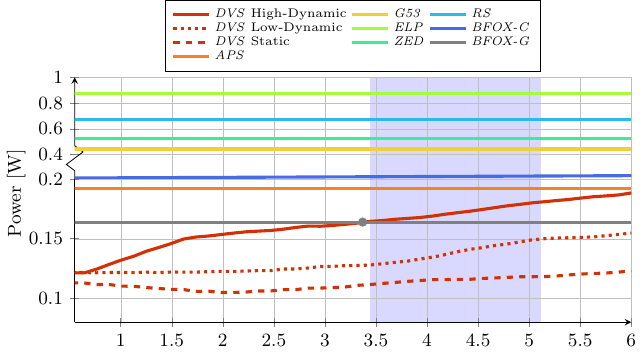}
    \includegraphics[width=0.8\linewidth,page=4]{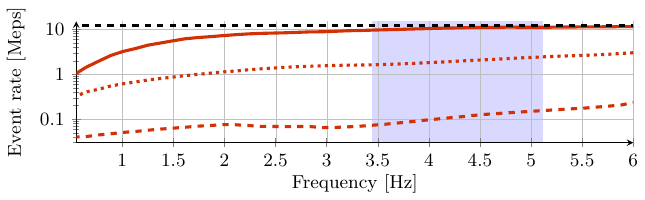}
    \caption{Top) Mean power consumed. Bottom) Mean number of events triggered by \textit{DVS}. The shadowed area corresponds to the flapping frequency range $\mu_{f} \pm \sigma_{f}$, see Section \ref{sec:application}.}
    \label{fig:power}
\end{figure}

Figure \ref{fig:power} shows the mean electrical power consumed by each camera as a function of the pitch rate. The presented values do not consider the standby power, which was experimentally measured when the cameras were not capturing information, before launching their drivers. Standby power is not reported as we aim at excluding embedded devices not involved in the event or image acquisition (e.g., LEDs, IMU, among others). The instantaneous pitch rate was obtained from the attitude measurements from the VectorNav using the Hilbert transform. The results showed that \textit{DVS} had lower consumption in this pitch range than most of the analyzed frame-based cameras. Only \textit{BFOX-G} camera had a lower consumption than \textit{DVS} in the \textit{high-dynamic scenario}. This result is relevant if we consider the consolidated technology of frame-based cameras versus the recent advent of event cameras, showing their low-energy consumption advantages.

\section{Ornithopter Flight Requirements}
\label{sec:flight}

\subsection{Dynamic Range}
Dynamic range quantifies the ability of a camera to capture different brightness levels in a scene. In outdoor scenarios, the ornithopters' pitch variations due to flapping strokes cause strong lighting conditions changes between brighter (e.g., sunlight) and darker (e.g., ground) illumination. In this experiment, we assess dynamic range by evaluating the capacity of detecting ArUco markers using data provided by the sensors under a wide range of illumination conditions. Each camera (previously calibrated) was installed inside a box pointing towards two 2$\times$2 ArUco boards (8 markers in total) attached to the back of the box. The boards were separated by a plate such that each one had different lighting, see Figure \ref{fig:setups}-c. The cameras were carefully set such that both boards were in its FoV. One board was illuminated with constant lighting $L_c = $ \SI{3}{\kilo\lux}, while the second one with increasing lighting $L_d(t)$ between $\sim$\SI{0}{\lux} and \SI{3}{\kilo\lux}. A luxmeter installed inside the box was used to register the lighting conditions. The lighting difference at time $t$ was computed as $20 \log_{10}(\frac{L_c}{L_d(t)})$. \textit{DVS} was moved slowly to enable event generation, while the rest remained static. Events were accumulated at \SI{30}{\hertz} to reconstruct frames using E2VID \cite{rebecq2021high}. ArUcos were detected with the method in \cite{garrido2014automatic} for all intensity images, including reconstructed frames. Frame cameras were set with autoexposure on, but limiting them to keep a minimum of 30 FPS, the lowest rate of all the analyzed cameras. This bound is selected to ensure a minimum reactivity, which is a critical requirement.

\begin{figure}
    \centering
    \includegraphics[width=0.8\linewidth,page=7]{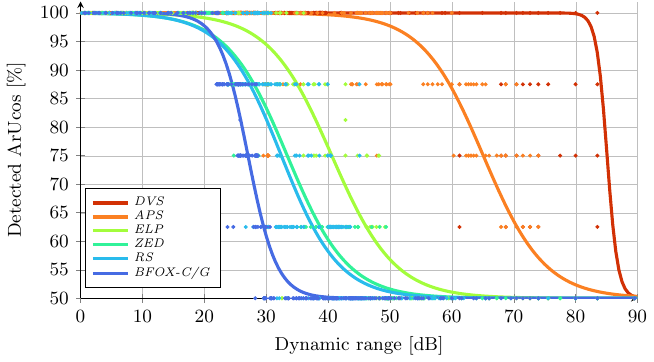}
    \caption{Dynamic range of the cameras in Table \ref{tab:cameras}. Points correspond to real measurements. Lines, computed by fitting the data using a sigmoid model, are depicted for visualization.}
    \label{fig:arucos}
\end{figure}

Figure \ref{fig:arucos} shows the percentage of the markers detected by each camera. The \textit{G53} was not considered as it includes a band-pass filter at \SI{850} {\nano\metre} (infrared light), which is out of the scope. The ArUco detection percentage always kept $>$50\% during the whole experiment as all cameras were able to detect the markers that were illuminated with $L_c$. \textit{DVS} reported the highest performance by detecting almost 100\% of the ArUcos in the range between \SI{0}{\decibel} to $\sim$\SI{83}{\decibel}. However, its detection performance decayed under pitch dark conditions (between \SI{83}{\decibel} and \SI{90}{\decibel}) due to the insufficient number of events and the large number of noise events, both hampering the reconstruction. Although the dynamic range specifications of \textit{DVS} is $>$\SI{120}{\decibel}, our experiment measured the ArUco detection performance under different lighting conditions. Additionally, this range also depends on the selected reference $L_c$. Further, \textit{APS} showed the best results among all frame cameras by detecting the 100\% of ArUcos until $\sim$\SI{45}{\decibel}. The results evidence the \textit{DVS} high dynamic range compared to all the frame cameras.

\label{subsec:dynamic}

\subsection{Motion Blur}
\label{subsec:motion_blur}
\begin{figure}
    \centering
    \includegraphics[trim={3em 5em 3em 0}, clip,width=0.32\linewidth,height=3.5cm]{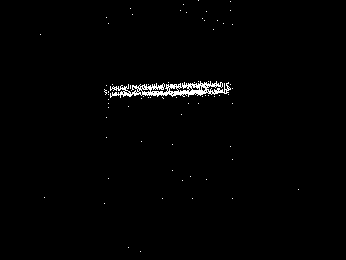}
    \includegraphics[trim={3em 5em 3em 0}, clip,width=0.32\linewidth,height=3.5cm]{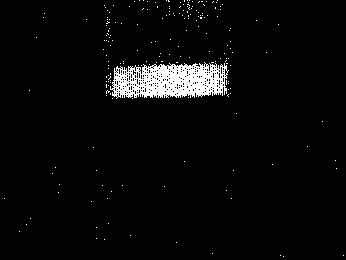}
    \includegraphics[trim={3em 5em 3em 0}, clip,width=0.32\linewidth,height=3.5cm]{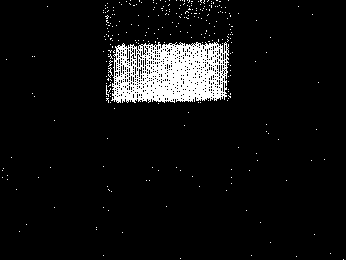}
    \caption{Motion blur and event cameras. From left to right, \textit{event images} integrating 1000, 5000, and 10000 events.}
    \label{fig:motion_blur_eimages}
\end{figure}

\begin{figure}
    \centering
    \includegraphics[width=0.8\linewidth,page=8]{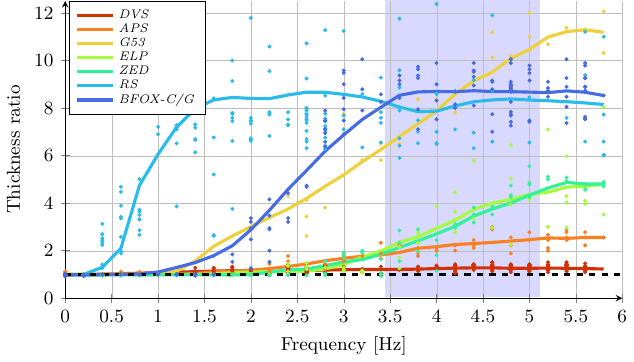}
    \caption{Motion blur (as the ratio between the thickness of blurred lines and their original thickness) versus camera pitch rate. Points correspond to real measurements while lines are depicted for visualization. The shadowed region corresponds to the flapping frequency range $\mu_{f} \pm \sigma_{f}$, see Section \ref{sec:application}.}
    \label{fig:motion_blur}
\end{figure}

To analyze the robustness against motion blur, each camera was set pointing to a horizontal white line placed over a black background. The setup was arranged with a constant illumination of \SI{800}{\lux}. The cameras were calibrated and mounted on the mechanism shown in Figure \ref{fig:setups}-b, which performed an oscillating movement (similar to flapping strokes) perpendicular to the line, hence maximizing the influence of motion blur. We evaluated the blur produced by a line as a function of the camera pitch rate --calculated as in Section \ref{subsec:power}. The events from \textit{DVS} were accumulated in \textit{event images} of 1000 events, experimentally selected to allow a proper definition of the line. Notice how a high number of events per frame can lead to images that integrate longer times, a phenomenon similiar to motion blur in standard images, see Fig. \ref{fig:motion_blur_eimages}. Frames with motion blur tend to describe blurred lines with irregular thickness. The line in all images, including the \textit{event images}, was detected with the image thresholding method in \cite{bradley2007adaptive}. The amount of blur was evaluated as the ratio between the thickness of each blurred line and its original thickness, computed when the camera remained almost static. This ratio was selected to ensure that the evaluation remains independent of the frame resolution. \textit{DVS} used the same reference as \textit{APS} as both sensors share the same pixel array. The results are presented in Figure \ref{fig:motion_blur}. As expected, the motion blur of frame cameras increased with the pitch rate. For \textit{DVS}, the thickness ratio remained close to one for all rates, evidencing high robustness. Although evaluating motion blur on \textit{event images} is not equivalent to use \textit{single events}, for fairness the adopted approach analyzed blur for all cameras using the same representation.

\renewcommand{\arraystretch}{1.5}
\setlength{\tabcolsep}{0.4em}
\begin{table*}
\begin{center}
\scalebox{0.8}{
\begin{tabular}{c|cccc|cccc|cccc|ccccc|}
&
\multicolumn{4}{c|}{\cellcolor{blue!50}{\textbf{Corner detection}}}
&
\multicolumn{4}{c|}{\cellcolor{blue!50}{\textbf{VIO}}}
&
\multicolumn{4}{c|}{\cellcolor{blue!50}{\textbf{Line detection}}}
&
\multicolumn{5}{c|}{\cellcolor{blue!50}{\textbf{Human detection}}}
\\
&
\cellcolor{blue!25}{\textbf{Pr.}} & \cellcolor{blue!25}{\textbf{Rec.}} & \cellcolor{blue!25}{\textbf{F1}} & \cellcolor{blue!25}{\textbf{Freq.}}
&
\cellcolor{blue!25}{\textbf{E$_{\mathrm{RMS}}$}} & \cellcolor{blue!25}{\textbf{E$_\sigma$}} & \cellcolor{blue!25}{\textbf{E$_{\mathrm{goal}}$}} & \cellcolor{blue!25}{\textbf{$\boldsymbol\varepsilon_{\mathrm{goal}}$}}
&
\cellcolor{blue!25}{\textbf{E$_\rho$}} & \cellcolor{blue!25}{\textbf{Prev.}} & \cellcolor{blue!25}{\textbf{N}} & \cellcolor{blue!25}{\textbf{Freq.}}
&
\cellcolor{blue!25}{\textbf{Acc.}} & \cellcolor{blue!25}{\textbf{Pr.}} & \cellcolor{blue!25}{\textbf{Rec.}} & \cellcolor{blue!25}{\textbf{F1}} & \cellcolor{blue!25}{\textbf{Freq.}}
\\
\textbf{\textit{APS}}
&0.972 & 0.806 & 0.881 & 39.92
&\textbf{0.828} & \textbf{0.439} & \textbf{1.804} & \textbf{0.231}
& 0.529 & \textbf{0.943} & 1.886 & 39.92
&\textbf{0.960} & 0.960 & \textbf{1.000} & \textbf{0.979} & 39.92
\\
\textbf{\textit{DVS}}
& 0.891 & \textbf{0.909} & \textbf{0.900} & \textbf{0.97M}
& 0.879 & 0.607 & 2.162 & 0.277
& 0.377 & 0.825 & \textbf{1.893} & \textbf{119.903}
& 0.909 & \textbf{0.985} & 0.921 & 0.952 & \textbf{57.69}
\\
\textbf{\textit{RS}}
& \textbf{1.000} & 0.014 & 0.028 & 29.96
& 1.178 & 0.698 & 2.369 & 0.304
& \textbf{0.255} & 0.234 & 0.659 & 29.96
& 0.759 & 0.869 & 0.857 & 0.863 & 29.96
\\
\end{tabular}
}
\caption{Evaluation results of different frame- and event-based algorithms. The performance metrics for \textit{corner detection} are \textit{Precision} (Pr.), \textit{Recall} (Rec.), \textit{F1-score} (F1), and the frame frequency (Freq., in Hertz). For \textit{single events}, Freq. corresponds to the mean number of events received per second. \textit{Visual inertial odometry} metrics are the root mean square error of the position (E$_{\mathrm{RMS}}$, in meters), the standard deviation of the position error (E$_\sigma$, in meters), the position error at the end of the trajectory (E$_{\mathrm{goal}}$, in meters), and the normalized final error ($\varepsilon_{\mathrm{goal}}$, i.e., E$_{\mathrm{goal}}$ divided by the total flown distance). The metrics for \textit{line detection} are the mean error (E$_\rho$), \textit{Prevalence} (Prev.), the mean number of detected lines per images ($N$), and the frame frequency (Freq., in Hertz). \textit{Human detection} metrics are \textit{Accuracy} (Acc.), Pr., Rec., F1, and Freq..}
\label{tab:results}
\end{center}
\end{table*}

\section{Application Dependent Requirements}
\label{sec:application}
\begin{figure}
    \centering
    \includegraphics[width=0.8\linewidth,page=9]{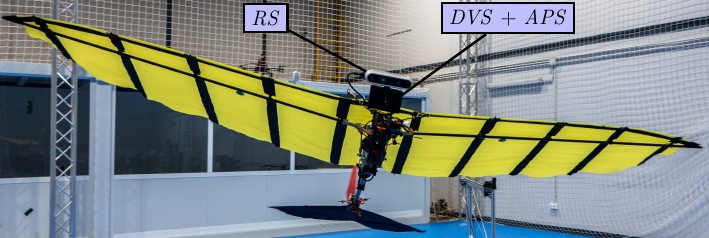}
    \caption{\textit{E-Flap} ornithopter during a data collection flight.}
    \label{fig:ornithopter}
\end{figure}

This section compares the performance of different vision algorithms by processing events and frames collected onboard the \textit{E-Flap} ornithopter \cite{zufferey2021design}; see Figure \ref{fig:ornithopter}. We registered measurements from a DAVIS346 (\textit{APS} and \textit{DVS}), an Intel RealSense D435 (\textit{RS}), and a VectorNav VN-200 inertial navigation system onboard the ornithopter. We selected the \textit{RS} as it is the most widely-used RGB sensor for robot perception among all evaluated cameras. The robot was controlled through a Khadas VIM3 that also recorded the sensor measurements. For a wider validation, we selected algorithms that input different representations: \textit{single events}, \textit{event images}, and \textit{reconstructed frames}, evidencing advantages and disadvantages of each representation versus traditional frames. The collected data were processed using publicly-available event-based and framed-based algorithms for \textit{corner detection}, \textit{visual-inertial odometry}, \textit{line detection}, and \textit{human detection}. It is worth mentioning that these experiments do not consider the processing onboard the ornithopter. Instead, they focus on validating the use of events and frames recorded to solve well-known perception problems. The experiments were performed in the GRVC Robotics Lab indoor testbed (\SI{15}{\meter}$\times$\SI{21}{\meter}$\times$\SI{8}{\meter}) equipped with 24 Optitrack cameras that provided the robot ground truth pose. In each experiment, the ornithopter flew describing a straight trajectory over different targets: i) a 7$\times$8 checkerboard, ii) a pattern with two horizontal lines, and iii) two people. More than 15 flights were performed. The robot traveled a mean distance of \SI{17}{\meter} with a mean velocity of \SI{3.06}{\meter / \second}. The ornithopter flew with a mean flapping rate $\mu_f = $ \SI{4.279}{\hertz} and standard deviation $\sigma_f = $ \SI{0.833}{\hertz}.

\subsection{Corner Detection}
Harris corner detector and its event-based version eHarris* \cite{vasco2017independent}\cite{alzugaray2018asynchronous} were used to detect the corners of a checkerboard. eHarris* processes \textit{single events} exploiting the high temporal resolution of event cameras. Manually annotated corners were used as ground truth. Ground truth corners were obtained by geometrically computing their locations using the 4 external corners position, the dimension and number of squares in the pattern, and the checkerboard size. We only considered the corners detected by eHarris* in a time window of \SI{10}{\milli\second} before the timestamp of each \textit{APS} image. A \textit{True Positive} occurred when a corner lied within a radius of \SI{3.5}{px} from the corner ground truth. Otherwise, it was considered a \textit{False Positive}. \textit{False Negatives} corresponded to ground truth references without any detected corner within a distance of \SI{3.5}{px}. \textit{Recall}, \textit{Precision}, and \textit{F1} were computed to evaluate the algorithms' performance. Frames reported a much lower number of corners compared to the rest of the pixels, and the number of non-corner events drastically varied depending on the camera motion and the scene dynamics. Thus, \textit{Accuracy} was not considered as the number of \textit{True Negatives} varied depending on the sensor. The results in Table \ref{tab:results} show that Harris with \textit{RS} had poor performance due to motion blur as evidenced by the low \textit{Recall}. Its high \textit{Precision} was caused by the low number of \textit{False Positives} as Harris did not detect corners in most cases. Conversely, Harris with \textit{APS} had a better performance. The low motion blur on \textit{APS}, see Section \ref{subsec:motion_blur}, favored detection. eHarris* reported the highest \textit{Recall} and \textit{F1}, although it showed lower \textit{Precision} than Harris with \textit{APS}. This was attributed to the larger number of \textit{False Positives} obtained with \textit{DVS} compared to those with \textit{APS}.

\subsection{Visual Odometry}
\begin{figure}
    \centering
    \includegraphics[page=5,width=0.8\linewidth,trim={1mm 1mm 0 0},clip]{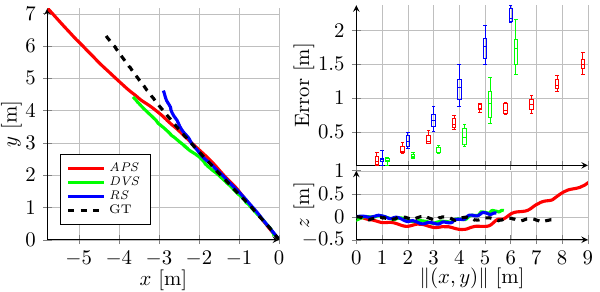}
    \caption{\textit{E-Flap} trajectory estimated with VINS-MONO (images) and Ultimate SLAM (events) in one flight.}
    \label{fig:vio}
\end{figure}

We evaluated two widely-known VIO pipelines: VINS-MONO \cite{qin2018vins} for images and Ultimate SLAM \cite{vidal2018ultimate} for events. The latter uses \textit{event images} instead of \textit{single events} to establish feature tracks. Both algorithms input IMU measurements from the VectorNav VN-200. Ground truth robot pose was measured with the Optitrack motion capture system. The root mean squared error E$_{\mathrm{RMS}}$, the standard deviation error E$_\sigma$, the final translation error E$_{\mathrm{goal}}$, and its normalized value $\boldsymbol\varepsilon_{\mathrm{goal}}$ were used as evaluation metrics. The average results obtained in all the flights are presented in Table \ref{tab:results}. The estimated trajectories in one flight are shown in Figure \ref{fig:vio}. VINS-MONO with \textit{APS} images reported the best results followed by Ultimate SLAM with \textit{DVS} \textit{event images}. VINS-MONO with \textit{RS} images described slightly worst E$_{\mathrm{RMS}}$ and E$_\sigma$ results as many RS images had motion blur. Although all methods reported acceptable E$_{\mathrm{RMS}}$ and E$_\sigma$, the final translation errors (\textbf{E$_{\mathrm{goal}}$} and \textbf{$\boldsymbol\varepsilon_{\mathrm{goal}}$}) were considerably large as the position estimations drifted along the robot trajectory. In fact, these errors were not larger due to the significantly accurate IMU data provided by the VN-200. Besides, the reported errors also depend on the performance of the evaluated algorithms. This evidences the current lack of VIO algorithms suitable for ornithopters.

\subsection{Line Detection}
We used the Hough transform algorithm, in which each line is represented by the tuple ($\rho$, $\theta$). The line detector input frames and \textit{event images} resulting from accumulating a constant number of events, experimentally set to 1250. We evaluated the detection of two horizontal lines located on one of the target boards. Lines not intersecting the target board or not satisfying $\| \theta \| \leq \SI{1}{\degree}$ --assuming ($\rho$, \SI{0}{\degree}) is an horizontal line-- were not considered for evaluation. The ground truth lines in \textit{APS} and \textit{RS} images were manually labeled. For \textit{event images}, they were obtained interpolating the annotations of the \textit{APS}. We used \textit{Prevalence} --positive detections divided by the total-- to define detection performance. If the number of correctly detected lines in one image is two, we consider it as a positive detection and, otherwise, as negative detection. Since $\| \theta \| \leq \SI{1}{\degree}$, we used E$_\rho$ to define the detection quality. E$_\rho$ corresponds to the mean distance between the detected and the ground truth lines in the $\rho$ axis of the Hough space. To compute E$_\rho$, the distances were normalized by the target board height in the image plane to make it independent of the robot-target distance. Table \ref{tab:results} shows the \textit{Prevalence}, E$_\rho$, and the mean number of lines detected $N$. \textit{APS} reported the best \textit{Prevalence}. The low motion blur experienced by \textit{APS} favored line detection. Conversely, motion blur drastically affected \textit{RS}, which reported the lowest \textit{Prevalence}. However, the positive line detections only corresponded to well-defined lines, favoring low E$_\rho$. The \textit{DVS} performance was very similar to that of \textit{APS} except for a small degradation when the lines were not well defined in the \textit{event images}.

\subsection{Human Detection}
Human detection was performed with YOLOv5 \cite{redmon2016you} using as input \textit{APS} and \textit{RS} images, and the \textit{reconstructed images} from events using E2VID \cite{rebecq2019events} with a fixed number of 3000 events. The ground truth was obtained by manually labeling with bounding boxes. For events, ground truth bounding boxes where interpolated. The difference between the detected and ground truth bounding boxes was estimated by the distance between the top-left corners of both boxes, and the differences between their heights and widths. If the sum of them three was lower than a experimentally set threshold, the detection was considered as a \textit{True Positive}. Otherwise, it was regarded as a \textit{False Positive}. Not detected people corresponded to \textit{False Negatives}, while images without people and detection as \textit{True Negatives}. To preserve the evaluation consistency, the results were computed using the same number of frames with and without people. The performance was evaluated with \textit{Accuracy}, \textit{Precision}, \textit{Recall}, and \textit{F1} metrics, see Table \ref{tab:results}. \textit{APS} had the best results, followed by \textit{DVS}. Although the method in \cite{rebecq2021high} generally provides reliable reconstruction, the flapping motion hampered the quality of the \textit{DVS} \textit{reconstructed images} compared to the \textit{APS} frames. Meanwhile, \textit{RS} reported the poorest performance due to motion blur.

\section{Discussion}
\label{sec:discussion}
In this section we intend to answer the question \textit{which vision sensor is more suitable for flapping-wing robots: frame or event cameras?}  In Sections \ref{sec:platform} and \ref{sec:flight}, the event camera outperformed the analyzed frame cameras in power consumption, robustness to motion blur, and dynamic range. Besides, current event cameras have similar sizes, weights, and resolutions to frame sensors. The analysis above suggests that event cameras are more convenient for ornithopters. Moreover, in Section \ref{sec:application}, we validated the use of events and frames recorded on board an ornithopter to solve perception problems. In this context, the results obtained with \textit{APS} frames were remarkable. Contrary to \textit{RS}, \textit{APS} was barely affected by the flapping strokes, see Section \ref{subsec:motion_blur}. Nonetheless, methods that input events provided similar performance. Our study suggests that the event representation has a key role in the performance. The results obtained using \textit{single events} and \textit{event images} tend to outperform those obtained with intensity frames. These event-based representations provided the fastest input rates, potentially allowing faster responses. Conversely, using \textit{reconstructed frames} from events collected on board ornithopters does not offer a significant advantage.

Despite our study suggesting event cameras as the most promising sensor for ornithopters, it does not consider the amount of available software resources. Frame-based methods are in a considerably maturer stage than event algorithms. There is a wide variety of traditional vision libraries and deep learning methods compared to those for events. This eases the implementation of frame-based methods. Nevertheless, considering the growing number of event algorithms \cite{gallego2020event} and the previous experimental results, this comparison suggests event-based vision as a promising solution for ornithopter perception. It is worth mentioning that our study only considers a low-resolution event camera. High-resolution event cameras offer similar dynamic range and temporal resolution. Besides, high-resolution event cameras report lower consumption than traditional cameras with similar resolutions \cite{finateu2020back}. Further, the study in \cite{gehrig2022are} suggests low-resolution event cameras perform better under high-speed motions and low illuminations. These are the conditions during flights, hence suggesting using low-resolution event cameras rather than others with higher resolution. Although our study addresses the main challenges for ornithopters, it lacks an analysis of the possible issues of processing events and frames onboard. These platforms typically mount lightweight computers with constrained computing power \cite{rodriguez2021griffin}, which is critical when the algorithms' latencies are higher than the sensor output rate. In this context, processing bottlenecks might occur using event cameras as they trigger more events under fast motions and dynamic scenes. In our experiments, the flapping-wing robot generated $\sim$0.97 million events per second, which is higher than the event-based corner detector rate ($\sim$0.22 million events per second \cite{alzugaray2018asynchronous}), evidencing processing bottlenecks. In fact, \cite{rodriguez2022free,tapia2020asap,tapia2022asap} integrate different approaches to filter and transmit events to prevent these issues. Nevertheless, frame methods may also suffer from bottlenecks on ornithopters. A future study is required to evaluate the processing issues of both events and frames on board ornithopters.

\section{Conclusions and Future Work}
\label{sec:conclusion}
The strict constraints and the challenging conditions that arise during flapping impose strong requirements for ornithopter perception. Previous works have explored the use of different vision sensors without having a consensus on which one is the most suitable. This paper addressed this question by experimentally evaluating several traditional cameras and an event camera considering the ornithopter flight conditions. We also validated the use of events and images on data captured on board a flapping-wing robot in widely-known computer vision tasks. Our results suggest event cameras as the most suitable vision sensor for ornithopters. Despite \textit{APS} and \textit{DVS} obtained similar results in the application dependent experiments, the current growth of the event vision community suggests future novel, faster, and more robust algorithms that exploit the advantages of event cameras for ornithopters. Future work focuses on extending the comparative onboard the robot to other frame cameras and analyzing the advantages and limitations of onboard processing events and frames on lightweight computers limited by resources.

\section*{Acknowledgements}
This work was funded by the European Research Council as part of GRIFFIN ERC Advanced Grant 2017 (Action 788247). Partial funding was obtained from the project ROBMIND (Ref. PDC2021-121524-I00) and from the Plan Estatal de Investigación Científica y Técnica y de Innovación of the Ministerio de Universidades del Gobierno de España (FPU19/04692 and FPU21/05333).

\bibliographystyle{IEEEtran}
\bibliography{iros2023}

\end{document}